\documentclass[conference]{IEEEtran}
\IEEEoverridecommandlockouts
\usepackage{cite}
\usepackage{amsmath,amssymb,amsfonts}
\usepackage{graphicx}
\usepackage{textcomp}
\usepackage{xcolor}
\usepackage{algorithm}
\usepackage{algpseudocode}
\usepackage{graphicx}
\usepackage{textcomp}
\usepackage{xcolor}
\usepackage{booktabs}
\usepackage{tabularx}
\usepackage{multirow}
\usepackage{array}
\newcolumntype{Y}{>{\centering\arraybackslash}X}

\usepackage{array}
\usepackage[normalem]{ulem} 
\usepackage{siunitx}
\usepackage{tcolorbox}
\usepackage{colortbl} 
\definecolor{graytxt}{gray}{0.55} 

\usepackage{amsmath}  
\usepackage{url}
\usepackage{amssymb}  
\usepackage[table]{xcolor}
\newcolumntype{O}{>{\columncolor{orange!20}}l}  
\newcolumntype{G}{>{\columncolor{green!20}}l}
\newcolumntype{R}{>{\columncolor{red!20}}l}
\newcolumntype{A}{>{\columncolor{gray!20}}l}
\newcolumntype{B}{>{\columncolor{blue!20}}l}
\def\BibTeX{{\rm B\kern-.05em{\sc i\kern-.025em b}\kern-.08em
    T\kern-.1667em\lower.7ex\hbox{E}\kern-.125emX}}
\begin{document}

\title{Leveraging LLM-Generated Explanations for Detecting Emotionally Rewritten Fake News}

\author{\IEEEauthorblockN{Yupei Guo}
\IEEEauthorblockA{{The University of Osaka}\\
Osaka, Japan \\
yupei@\\im.sanken.osaka-u.ac.jp}
\and
\IEEEauthorblockN{Jiajun He}
\IEEEauthorblockA{{Alibaba Inc}\\
Hangzhou, China \\
kakashun.hjj@alibaba-inc.com}
\and
\IEEEauthorblockN{Xiaohan Shi}
\IEEEauthorblockA{{Nagoya University}\\
Nagoya, Japan \\
xiaohan.shi@g.sp.m.is.nagoya-u.ac.jp}
\and
\IEEEauthorblockN{Tomoki Toda}
\IEEEauthorblockA{{Nagoya University} \\
Nagoya, Japan \\
toda.tomoki.v6@f.mail.nagoya-u.ac.jp}
\and
\IEEEauthorblockN{Zekun Yang}
\IEEEauthorblockA{{Tokyo University of Science}\\
Tokyo, Japan \\
yang.zekun@\\rs.tus.ac.jp}
\and
\IEEEauthorblockN{Bowen Wang}
\IEEEauthorblockA{{The University of Osaka}\\
Osaka, Japan \\
wang@\\im.sanken.osaka-u.ac.jp}
\and
\IEEEauthorblockN{Yukinobu Taniguchi}
\IEEEauthorblockA{{Tokyo University of Science}\\
Tokyo, Japan \\
taniguchi.yukinobu@\\rs.tus.ac.jp}
}

\maketitle

\begin{abstract}
The spread of fake news may cause severe social consequences. 
Existing fake news detection methods mainly focus on stylistic variations or incorporate external information such as explanations. However, news articles are often rewritten under different emotional backgrounds while preserving their underlying factual claims, which may affect the robustness of detection models. In this work, we investigate fake news detection under fact-preserving emotional variations. To study this problem, we construct emotion-rewritten test sets and generate explanations from the original news articles as stable background knowledge. We then propose a Gated Cross Attention (GCA) framework that adaptively integrates emotionally rewritten news with the corresponding explanations, enabling the model to focus on informative explanation content while reducing potential mismatches caused by emotional reframing. Experiments on PolitiFact, GossipCop, and LUN \textcolor{black}{demonstrate} that the proposed method \textcolor{black}{achieves notable improvements under multiple emotional conditions on PolitiFact and LUN, while maintaining competitive performance on GossipCop}. We further analyze the effects of explanation guidance and gating mechanisms under different emotional conditions. \textcolor{black}{Our code and data are available at: \url{https://github.com/Flulike/fakenews_gca} }.
\end{abstract}


\section{Introduction}
\label{sec:1}

News is an important medium for people to know what is happening. 
With the rapid development of social network platforms, a large amount of news is produced and disseminated every day. 
However, the open and decentralized nature of these platforms also facilitates the spread of false or misleading information, which can be called fake news~\cite{RN22,RN23}. 
Such fake news influences people's decisions, undermines social credibility, and may even cause severe consequences~\cite{RN13,RN24}. 
Moreover, it is often difficult for ordinary users to verify the authenticity of online content, allowing fake news gain traction~\cite{RN05,RN24}. 
Therefore, developing accurate fake news detection systems remains an important research problem.

\begin{figure}[t]
  \centering
  \includegraphics[width=.5\textwidth]{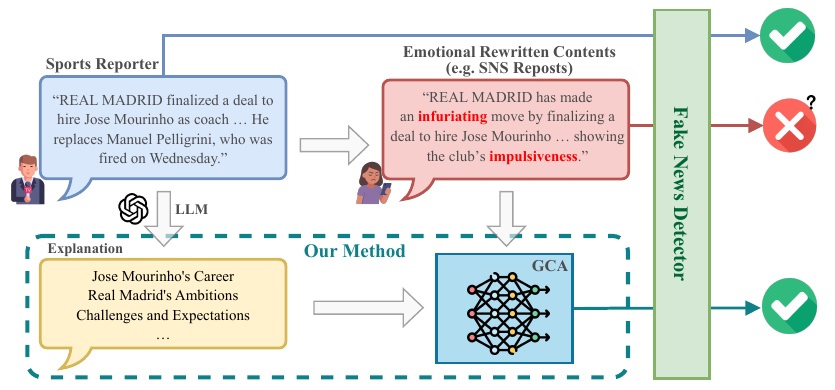}
  \caption{Overview of fake news detection framework under different emotional backgrounds. Red text indicates emotion-related modifications introduced during rewriting.}
  \label{fig:1}
\end{figure}

Recent research on automated fake news detection mainly follows two complementary directions: studying the influence of style variations~\cite{RN08,RN09,RN04,RN11,RN18,RN27,wang2026megafake,tahmasebi2026robust,wu2027opend}, and leveraging external information as supplementary evidence~\cite{RN01,RN02,RN10,RN12,RN03,RN14,RN19,lu2026llm,he2026factguard}. For example, Przybyla et al.~\cite{RN08} detect fake news based on stylistic characteristics, while Wu et al.~\cite{RN04} propose a style-robust detector that emphasizes content-related features. \textcolor{black}{Wang et al.~\cite{wang2026megafake} introduce MegaFake, a theory-driven dataset constructed by simulating socio-psychological style variations and deceptive mechanisms in LLM-generated fake news.} Xie et al.~\cite{RN01} combine knowledge graphs, large language models, and graph neural networks; Hu et al.~\cite{RN02} design an Adaptive Rationale Guidance (ARG) network; and Nan et al.~\cite{RN03} augment fake news detection with generated comments. Ma et al.~\cite{RN28} further demonstrate that extracting structured semantic information such as entities and topics with large language models can improve fake news detection compared with directly using LLM embeddings. \textcolor{black}{Lu et al.~\cite{lu2026llm} propose a method that  simulate user discussions to alleviate data scarcity.}  

Other studies have also demonstrated that fake news often leverages emotionally charged language to influence readers. For instance, fabricators of false information deliberately employ specific emotional tones to mask unverified content and accelerate dissemination~\cite{liu2024emotion}. Consequently, incorporating emotional signals has become an important research direction in fake news detection. Recent studies have explored advanced emotional representations that capture nuanced publisher and social emotions as complementary features~\cite{toughrai2025fake}. However, treating emotion primarily as a detection cue may introduce a potential vulnerability. In real-world scenarios, news articles are often rewritten under different emotional backgrounds to reflect certain stances or tendencies while largely preserving their underlying claims. As a result, detectors that rely heavily on emotional cues may be misled by altered emotional expressions. Such emotional perturbations make it more difficult to separate emotional presentation from factual content, posing challenges to fake news detection even when auxiliary information is available. Therefore, it is necessary to develop methods that robustly utilize external evidence while remaining resilient to changes under different emotional backgrounds.

In this work, we investigate fake news detection under fact-preserving emotional variations, where the underlying factual content remains unchanged while the emotional background of the news varies. The setting is shown in Fig.~\ref{fig:1}. Following prior work on style-preserving news rewriting~\cite{RN04}, we adopt this setting to isolate the influence of emotional variations from changes in factual content. To study this problem, we construct emotionally rewritten versions of news articles and pair them with explanations generated from the corresponding original articles. These explanations serve as stable background knowledge across different emotional variants. Although such explanations may not always be available in real-world deployment scenarios, they are used here as a reference to examine how auxiliary information interacts with emotionally altered inputs. This design enables us to analyze whether fake news detectors can maintain consistent judgments when exposed to different emotional expressions of the same content.


Based on this setting, we propose a Gated Cross Attention (GCA) framework to integrate emotionally rewritten news with explanation signals. \textcolor{black}{Unlike SheepDog~\cite{RN04}, which uses style-preserving rewriting to improve robustness without incorporating auxiliary explanations, our method introduces explanations generated from the original news as contextual information for emotionally rewritten inputs. Unlike ARG~\cite{RN02}, which uses rationales for general fake news detection, our method further models the varying relevance between an explanation and its emotionally rewritten news through gated cross attention. Specifically, the gate adaptively} regulates the contribution of text explanation according to the input representation, rather than treating the explanation as uniformly relevant. This provides a methodological extension from explanation augmentation to adaptive explanation-guided detection under emotional rewriting. 

We evaluate our approach on PolitiFact, GossipCop, and LUN. The proposed method achieves improvements under multiple emotional conditions on PolitiFact and LUN while remaining competitive on GossipCop. Ablation studies further examine the contributions of explanation augmentation and the gating mechanism.

\section{Dataset Creation}
\label{sec:2}
\subsection{Data Source}
We follow SheepDog~\cite{RN04} and use three widely adopted datasets: PolitiFact~\cite{RN06}, GossipCop~\cite{RN06}, and LUN~\cite{RN07}. PolitiFact and LUN mainly contain political news, whereas GossipCop focuses on entertainment and celebrity-related news. Following SheepDog, all datasets are formulated as binary fake news detection tasks with 0 for true news and 1 for fake news.

\begin{table}[t]
\centering
\caption{Statistics of the benchmark datasets used in this work. The three datasets cover different data scales, providing a comprehensive evaluation under multiple emotional rewriting settings.}
\label{tab:3}
\setlength{\tabcolsep}{6pt}
\renewcommand{\arraystretch}{1.1}
\small
\begin{tabular}{lcccccc}
\toprule
\textbf{Dataset} &
\multicolumn{3}{c}{\textbf{Training}} &
\multicolumn{3}{c}{\textbf{Test}} \\
\cmidrule(lr){2-4}
\cmidrule(lr){5-7}
& Total & True & Fake &
Total & True & Fake \\
\midrule
PolitiFact & 360 & 180 & 180 & 90 & 45 & 45 \\
GossipCop  & 6,332 & 3,166 & 3,166 & 1,584 & 792 & 792 \\
LUN        & 6,000 & 3,000 & 3,000 & 1,500 & 750 & 750 \\
\bottomrule
\end{tabular}
\end{table}

Table~\ref{tab:3} summarizes the original train/test split of the three datasets. The datasets cover both small-scale (PolitiFact) and large-scale (GossipCop and LUN) settings. For all datasets, the numbers of true and fake news are balanced in both the training and test sets, minimizing potential bias caused by class imbalance and enabling a fair comparison across different emotional rewriting settings. During training, each original news article is accompanied by style-preserving rewritings. Specifically, objective and professional and neutral rewritings are treated as reliable variants, while emotionally triggering and sensational rewritings are regarded as unreliable variants, following the setting of SheepDog~\cite{RN04}.

\subsection{Explanation Generation}
In real-world news consumption, authenticity judgments are often supported by background knowledge beyond the news article itself. To simulate such contextual information in a controlled manner, we employ GPT-4o~\cite{RN15} to generate explanations based on the original news text.

For each article, we use the following prompt:

\begin{center}
\fbox{
\begin{minipage}{0.9\columnwidth}
\textbf{Prompt for Explanation Generation} \vspace{1.5mm}
\\
Please give extra information to the following article:
[news article]
\end{minipage}
}
\end{center}

The generated explanations provide complementary semantic context related to the news content and are incorporated as auxiliary information during our detection. Since all emotionally rewritten variants originate from the same source article, re-annotating the authenticity of every rewritten variant is generally impractical, as the rewritten articles do not naturally occur in real-world datasets and no independent ground-truth labels are available. Following our fact-preserving rewriting assumption, the rewritten variants therefore inherit the authenticity label of the original article. Consequently, explanations are generated only once from the original article and reused for all rewritten variants. This design ensures that all variants share identical contextual information, allowing the experiments to isolate the influence of emotional expression while minimizing additional factors from repeated explanation generation.

It should be noted that these explanations are not intended to serve as independently verified external evidence. Instead, they function as contextual priors for evaluating robustness under controlled fact-preserving emotional rewriting. \textcolor{black}{The use of GPT-4o is intended to provide a controlled and consistent way of constructing such priors, rather than to prescribe LLM-based explanation generation as a necessary deployment strategy.} This setting enables us to investigate whether fake news detection models can maintain consistent predictions when only the emotional framing changes while the underlying semantic content remains unchanged.

\subsection{Test Set Emotional Rewritings}
Recent work explores robustness-oriented fake news detection through style rewriting while preserving the authenticity labels of the original articles~\cite{RN04}. Following this setting, we construct emotion-based rewritings to investigate the influence of emotional framing on fake news detection.

Specifically, we consider four commonly observed emotions: \textit{Joy}, \textit{Anger}, \textit{Sadness}, and \textit{Happiness}. GPT-4o~\cite{RN15} is employed to rewrite each test article using the prompt:

\begin{center}
\fbox{
\begin{minipage}{0.9\columnwidth}
\textbf{Prompt for Emotional Rewriting}
\vspace{1.5mm} \\
Rewrite the following article using the [emotion class] emotion: [news article]
\end{minipage}
}
\end{center}

Note that explanations and different emotional rewritings are generated independently through separate GPT-4o queries. Both procedures only used the original news article as input and did not access each other's outputs.

Following previous work~\cite{RN04}, we assume that emotionally rewritten articles inherit the authenticity label of the original news article. This assumption enables us to isolate the influence of emotional expression while keeping the underlying factual content unchanged.

For the training set, we retain the original setting of~\cite{RN04}, which contains the original articles together with reliable and unreliable style rewritings. In contrast, the test set is fully reconstructed using emotion-based rewritings. Therefore, the reliable and unreliable rewritings in training are independent of the emotion-based rewritings in evaluation.

To ensure the quality of the generated data, we manually inspected a subset of the rewritten articles and verified that the major factual claims were generally preserved while the emotional expressions were modified.

We note that this label inheritance assumption may not fully capture real-world situations where emotional framing and factual content change simultaneously. Nevertheless, LLM-based rewriting provides a controllable approximation that enables systematic evaluation of robustness under different emotional backgrounds. More realistic scenarios involving human-authored rewritings and simultaneous factual modifications are left for future work.



\begin{figure*}[t]
  \centering
  \includegraphics[width=.95\textwidth]{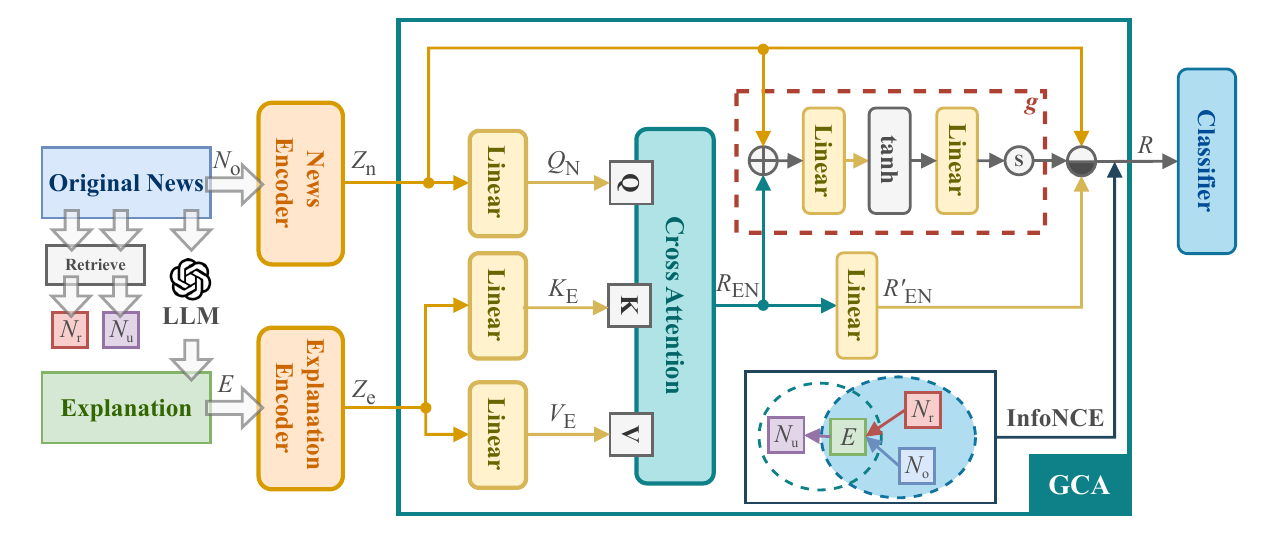}
  \caption{Overview of the proposed Gated Cross Attention (GCA) framework. The original article, one reliable rewriting, one unreliable rewriting, and explanation generated from the original article are encoded independently. GCA adaptively integrates explanation-aware contextual information to detect fake news.}
  \label{fig:2}
\end{figure*}


\section{Proposed Method}
\label{sec:3}
We propose a Gated Cross Attention framework (GCA) that integrates news content with LLM-generated explanations for fake news detection, as illustrated in Fig.~\ref{fig:2}. Our method extends Sheepdog~\cite{RN04} by incorporating explanations as contextual priors derived from the original news article. Specifically, following Sheepdog, each news piece is associated with three style-based variants: the original article, one reliable rewriting, and one unreliable rewriting. These variants share the same underlying content and authenticity label while differing in emotional and stylistic expressions, enabling controlled evaluation of robustness against emotional reframing.

In addition to these news variants, we associate each article with an explanation generated from the original news article. The explanation provides topic-related contextual information that may support authenticity judgment beyond surface-level expressions. However, directly combining explanation and news representations may introduce irrelevant or inconsistent information. Therefore, we design a gated cross-attention module to adaptively control the contribution of explanation features according to their relevance to the current news representation.

Specifically, we employ four pre-trained language models with identical architectures but independent parameters to encode the three news variants and the explanation, respectively. The gated cross-attention module then performs information interaction between news and explanation representations, where the learned gate dynamically adjusts the amount of explanation-aware information incorporated into the final representation. Finally, the fused representation is used to predict the authenticity label of the news article.



\subsection{News Rewritings and Explanation} 
Following Sheepdog~\cite{RN04}, we use three versions of each news article: the original article $N_o$, one reliable rewriting $N_r$, and one unreliable rewriting $N_u$. Unlike conventional retrieval systems, ${N}_{r}$ and ${N}_{u}$ are not external articles. Instead, they are stylistic rewritings of the same news piece released by Sheepdog and therefore share the same authenticity label as ${N}_{o}$. 

For each news article, we additionally use the generated explanation ${E}$ described in Section~\ref{sec:2}. The explanation is derived from the original news article and reused across all rewritten versions of the same news piece. It serves as a contextual prior that provides supplementary background information beyond the news text itself. This design enables us to examine whether explanation-based reasoning remains effective when the emotional framing of the news changes while the underlying content remains largely unchanged.

\subsection{GCA Fusion} 
To incorporate such contextual priors into news representations, we employ a cross-attention mechanism where the news representation acts as the query and the explanation representation serves as the key and value. 

Let ${Z}_{n}\in\mathbb{R}^{L\times d}$ denote the hidden representation of the news encoder and ${Z}_{e}\in\mathbb{R}^{L_e\times d}$ denote the hidden representation of the explanation encoder, where $L$ and $L_e$ are sequence lengths and $d$ is the hidden dimension. Linear projections are first applied with $W_Q$, $W_K$, and $W_V \in \mathbb{R}^{d\times d}$ being trainable projection matrices:
\begin{equation}
\begin{aligned}
Q_N = Z_n W_Q, K_E = Z_e W_K, V_E = Z_e W_V
\end{aligned}
\end{equation}

Cross-attention is then performed with ${R}_{EN}$ representing explanation-aware news features:

\begin{equation}
R_{EN}={\rm CA}(Q_N,K_E,V_E),
\end{equation}

Explanations are generated from the original article and reused across emotionally rewritings, their relevance may vary under different emotional contexts. Directly injecting explanation information may therefore introduce irrelevant or misleading information when the emotional rewriting changes the surface expression of the news. To mitigate this issue, we introduce a token-level gate ${g}\in\mathbb{R}^{L\times d}$ to dynamically regulate the contribution of explanation-aware features:

\begin{equation}
g=\sigma(G_{2}{\rm tanh}(G_{1}({Z}_{n}||{R}_{EN})+b_1)+b_2),
\end{equation}
where $||$ denotes concatenation, $\sigma$ is the element-wise sigmoid function, ${G}_{1}\in\mathbb{R}^{2d\times d}$ and ${G}_{2}\in\mathbb{R}^{d\times d}$ are trainable matrices, ${b}_{1}$ and ${b}_{2}$ are trainable bias terms.

To align feature dimensions, we apply a trainable projection where ${W}_{en}\in\mathbb{R}^{d\times d}$:

\begin{equation}
R'_{EN}=R_{EN}W_{en},
\end{equation}

The final fused representation is obtained by

\begin{equation}
R=g\odot R'_{EN}+(1-g)\odot Z_n,
\end{equation}
where $\odot$ denotes element-wise multiplication.

The fused representation is mean-pooled and conveyed into a dropout layer and a fully connected classifier to obtain the final authenticity prediction.

\subsection{Loss Function} 
In addition to the original training objectives of Sheepdog~\cite{RN04}, we introduce an InfoNCE~\cite{RN17} loss to improve the compatibility between news representations and source-derived explanations. Sequence-level representations are obtained by mean pooling over token embeddings, and the cosine similarity is computed between each explanation $E$ and its corresponding news representations ($N_o$, $N_r$, $N_u$), while similarities to other explanations in the same batch are minimized. The loss is weighted by the mean gate value of each news instance. Specifically,
\begin{equation}
\alpha_N=1-\operatorname{mean}(G_N),
\end{equation}
where $G_N$ denotes the token-level gate representation, and the gate is detached when computing $\alpha_N$. The contrastive loss is defined as
\begin{equation}
\begin{aligned}
L_{con}=\frac{1}{3}\big(&\alpha_o\operatorname{InfoNCE}(N_o,E)
+\alpha_r\operatorname{InfoNCE}(N_r,E)\\
&+\alpha_u\operatorname{InfoNCE}(N_u,E)\big).
\end{aligned}
\end{equation}
When both the gating module and $L_{con}$ are removed, the framework reduces to a straightforward explanation-enhanced variant of Sheepdog, as further examined in the ablation study.


\begin{table*}[t]
\centering
\caption{Experimental results of our method on the three benchmark datasets (in ${\%}$). Accuracy and F1 score are reported under the original news setting as well as different emotional rewriting conditions. }
\label{tab:1}
\setlength{\tabcolsep}{4pt}
\resizebox{\linewidth}{!}{%
\begin{tabular}{l|*{5}{c}|*{5}{c}|*{5}{c}}
\toprule
 & \multicolumn{5}{c|}{PolitiFact} & \multicolumn{5}{c|}{GossipCop} & \multicolumn{5}{c}{LUN} \\
Method
 & Orig & Joy & Anger & Sad & Happy
 & Orig & Joy & Anger & Sad & Happy
 & Orig & Joy & Anger & Sad & Happy \\
\midrule
\multicolumn{16}{B}{\textit{Accuracy}}    \\
ENDEF
 & 88.89 & 73.33 & 82.22 & 80.00 & 73.33
 & 67.99 & 63.13 & 68.50 & 66.35 & 61.81
 & 77.67 & 76.13 & 73.40 & 75.13 & 75.73 \\
ARG
 & 85.56 & 69.44 & 74.44 & 73.33 & 70.00
 & 73.23 & 69.67 & 66.79 & 68.75 & 69.88
 & 85.40 & 78.50 & 78.93 & 80.00 & 80.10 \\
Sheepdog
 & 88.77 & 76.11 & \textbf{84.22} & 79.89 & 71.67
 & 75.89 & 75.11 & \textbf{75.95} & 76.03 & 74.66
 & 92.61 & 85.90 & 88.43 & 86.71 & 87.41 \\
Ours
 & \textbf{90.56} & \textbf{77.11} & 83.56 & \textbf{81.11} & \textbf{75.56}
 & \textbf{76.72} & \textbf{76.29} & 75.38 & \textbf{76.10} & \textbf{76.18}
 & \textbf{93.96} & \textbf{87.29} & \textbf{90.01} & \textbf{88.20} & \textbf{88.55} \\
\midrule
\multicolumn{16}{R}{\textit{F1 Score}}    \\
ENDEF
 & 88.37 & 65.71 & 83.33 & 75.68 & 65.71
 & 61.33 & 46.72 & 64.18 & 57.46 & 42.55
 & 73.56 & 77.70 & 77.13 & 74.47 & 75.73 \\
ARG
 & 85.54 & 67.41 & 73.38 & 72.09 & 68.09
 & 73.13 & 69.67 & 66.35 & 68.66 & 69.88
 & 85.31 & 78.39 & 78.84 & 79.80 & 80.30 \\
Sheepdog
 & 88.74 & 75.66 & \textbf{84.17} & \textbf{79.64} & 70.20
 & 75.85 & 74.97 & \textbf{75.83} & 76.01 & 74.51
 & 92.59 & 85.87 & 88.40 & 86.67 & 87.41 \\
Ours
 & \textbf{90.51} & \textbf{76.14} & 83.36 & 79.41 & \textbf{74.70}
 & \textbf{76.28} & \textbf{76.27} & 75.05 & \textbf{76.03} & \textbf{76.14}
 & \textbf{93.95} & \textbf{87.27} & \textbf{90.01} & \textbf{88.15} & \textbf{88.53} \\
\bottomrule
\end{tabular}%
}
\end{table*}

\section{Experiments}
\label{sec:4}

\subsection{Experiment Setting} 
We conduct all experiments on a computer with AMD EPYC 7313 CPU, 128 GB memory, a 4 TB HDD, and an NVIDIA RTX A4000 GPU. Our implementation is based on Python 3.12 and PyTorch 2.5.1 under Ubuntu 20.04.

We use RoBERTa~\cite{RN16} as the backbone encoder and initialize four independent \texttt{roberta-base} models to encode the original news article, reliable rewriting, unreliable rewriting, and explanation, respectively. The maximum sequence length is set to 512 tokens for both news articles and explanations, with longer sequences being truncated. \textcolor{black}{We use AdamW as the optimizer with a learning rate of $2\times10^{-5}$, a weight decay of 0.01, and a batch size of 4. A linear learning-rate scheduler is applied with 0 warm-up steps and 10,000 total training steps. The models are trained for 5 epochs without early stopping.} The temperature parameter $\tau$ in the contrastive loss is set to 0.1. Following previous work~\cite{RN04}, we evaluate models on both the original test set and the emotion-rewritten test sets. \textcolor{black}{Each experiment is repeated 10 times using random seeds from 0 to 9, and} we report the mean Accuracy and F1 Score.

We compare our method with Sheepdog~\cite{RN04}, ENDEF~\cite{RN25}, and ARG~\cite{RN02}. Sheepdog is trained with the original text-only setting, while ENDEF uses both the news content and explanations. The ARG rationale inputs are replaced with explanations generated in Section~\ref{sec:2} to make use of external information. Experiments are conducted on PolitiFact~\cite{RN06}, GossipCop~\cite{RN06}, and LUN~\cite{RN07} datasets.

\begin{table*}[t]
\centering
\caption{Ablation study results of our method on three datasets (in ${\%}$). Note that removing the gating mechanism also disables $L_{con}$, since the contrastive loss is computed from gated representations.}
\label{tab:2}
\setlength{\tabcolsep}{4pt}
\resizebox{\linewidth}{!}{%
\begin{tabular}{l|*{4}{c}|*{4}{c}|*{4}{c}}
\toprule
 & \multicolumn{4}{c|}{PolitiFact} & \multicolumn{4}{c|}{GossipCop} & \multicolumn{4}{c}{LUN} \\
Method
 & Joy & Anger & Sad & Happy
 & Joy & Anger & Sad & Happy
 & Joy & Anger & Sad & Happy \\
\midrule
\multicolumn{13}{B}{\textit{Accuracy}}    \\
with BERT
 & 76.56 & 80.11 & 79.44 & 75.44
 & 75.42 & 75.04 & 75.70 & 74.99
 & 85.20 & 85.61 & 86.07 & 85.49 \\
w/o ${L}_{con}$
 & 76.11 & 82.29 & 81.11 & 75.05
 & 74.85 & \textbf{75.68} & 75.87 & 74.99
 & 86.40 & 89.93 & 88.10 & 88.33 \\
w/o Gate
 & \textbf{77.78} & 80.00 & 76.22 & 73.33
 & 75.51 & 75.63 & 75.82 & 76.12
 & 83.27 & 84.93 & 86.14 & 84.27 \\
\midrule
Ours
 & 77.11 & \textbf{83.56} & \textbf{80.11} & \textbf{75.56}
 & \textbf{76.29} & 75.38 & \textbf{76.10} & \textbf{76.18}
 & \textbf{87.29} & \textbf{90.01} & \textbf{88.20} & \textbf{88.55} \\
\midrule
\multicolumn{13}{R}{\textit{F1 Score}}    \\
with BERT
 & 76.32 & 80.08 & 79.32 & 76.05
 & 75.27 & 74.89 & 75.65 & 74.82
 & 85.12 & 85.52 & 86.03 & 86.24 \\
w/o ${L}_{con}$
 & 75.99 & 82.33 & \textbf{81.07} & 74.63
 & 74.75 & \textbf{75.61} & 75.61 & 74.86
 & 86.39 & 89.93 & 88.08 & 88.32 \\
w/o Gate
 & \textbf{77.78} & 79.96 & 76.19 & 73.32
 & 75.49 & 75.60 & 75.90 & 71.18
 & 83.15 & 84.77 & 85.99 & 84.27 \\
\midrule
Ours
 & 76.14 & \textbf{83.36} & 79.41 & \textbf{74.70}
 & \textbf{76.27} & 75.05 & \textbf{76.03} & \textbf{76.14}
 & \textbf{87.27} & \textbf{90.01} & \textbf{88.15} & \textbf{88.53} \\
\bottomrule
\end{tabular}
}
\end{table*}






\subsection{Experimental Results}


We evaluate our method using accuracy and F1 score on both the original news articles and their emotionally rewritten variants. The results are reported in Table~\ref{tab:1}. Overall, the proposed GCA framework improves fake news detection performance on PolitiFact and LUN while maintaining competitive performance on GossipCop under different emotional backgrounds. \textcolor{black}{The magnitude of the performance change after emotional rewriting varies substantially across datasets and emotional conditions, with relatively small changes on GossipCop and more pronounced changes on PolitiFact and LUN.} These results suggest that explanation-based contextual priors can improve robustness against emotional reframing without requiring modifications to the rewritten text.

On \textbf{PolitiFact}, our method achieves the best performance on the original news set as well as most emotionally rewritten variants. The improvements are consistently observed across different emotional conditions, suggesting that the generated explanations provide complementary semantic context that remains useful even when the surface expressions of the news articles change. This indicates that the proposed gated fusion mechanism can effectively utilize contextual information while suppressing less relevant explanation features.

On \textbf{LUN}, our method obtains the best overall performance among all evaluated settings. Compared with PolitiFact, the performance improvements are generally more pronounced. One possible explanation is that LUN mainly contains political news with relatively stable factual structures, making the generated contextual explanations more semantically consistent with the rewritten articles. Consequently, explanation-aware fusion becomes more effective under controlled emotional perturbations.

For \textbf{GossipCop}, the performance improvements are relatively modest. GossipCop consists primarily of entertainment and celebrity news, where the original articles already contain stronger emotional expressions and richer stylistic variations. As a result, emotional rewritings introduce comparatively smaller semantic changes, reducing the additional benefit provided by explanation-based contextual information. Nevertheless, our method remains competitive across all evaluated emotional backgrounds without causing noticeable performance degradation.

Another observation is that the proposed framework performs consistently across different emotional categories rather than improving only a specific rewriting style. This suggests that the explanation representations capture relatively stable semantic information that is less sensitive to stylistic variation, allowing the gating mechanism to selectively incorporate useful contextual knowledge during prediction.

Overall, these results demonstrate that integrating explanation-based contextual priors through the proposed GCA framework improves fake news detection under controlled emotional reframing. Rather than relying solely on surface textual features, the model benefits from stable semantic context provided by explanations, leading to more robust predictions across different emotional backgrounds.

\subsection{Ablation Studies}
To investigate the contribution of different components, we conduct ablation studies on the backbone encoder, the contrastive loss ${L}_{con}$, and the gating mechanism ${g}$. \textcolor{black}{Note that these components are not fully independent in our framework because $L_{con}$ is computed from gated representations.} Specifically, we replace RoBERTa with BERT, remove ${L}_{con}$, and remove the gating module. Since ${L}_{con}$ is applied on gated representations, removing ${g}$ also disables ${L}_{con}$. The results are summarized in Table~\ref{tab:2}.

The model without the gating mechanism and contrastive loss can be regarded as a simplified explanation-enhanced baseline, where the generated explanations are directly fused with the news representations without adaptive filtering. \textcolor{black}{This setting provides a reference for evaluating whether the full GCA framework offers benefits beyond simply introducing LLM-generated explanations.}

\textbf{Effect of the Backbone Encoder:} Replacing RoBERTa with BERT generally reduces performance across datasets and emotional backgrounds. This observation suggests that stronger contextual representations facilitate more effective semantic interaction between news articles and explanations, leading to more reliable explanation-aware fusion.

\textbf{Effect of the Contrastive Loss ${L}_{con}$:} Removing ${L}_{con}$ results in relatively modest but consistent performance degradation. Although the influence is smaller than that of the gating mechanism, the overall results indicate that contrastive learning improves the alignment between news articles and their corresponding explanations, making the fused representations more robust under different emotional rewritings.

\textbf{Effect of the Gating Mechanism:} 
The gate and $L_{con}$ are coupled through the loss weighting: $L_{con}$ is computed from pre-fusion representations, while its contribution is modulated by a gate-dependent coefficient. Removing the gating mechanism\textcolor{black}{, which also removes \(L_{con}\),} causes the largest performance degradation, particularly on LUN, and this trend remains consistent across most emotional backgrounds. \textcolor{black}{The comparison therefore evaluates the combined effect of adaptive gating and the associated contrastive alignment objective, rather than the gating mechanism in complete isolation.} These results indicate that directly incorporating explanation representations without adaptive filtering can introduce irrelevant information into the final representation. \textcolor{black}{The gating mechanism helps regulate the contribution of explanation-aware features according to their relevance to the current news representation, thereby reducing such interference.}

Compared with the explanation-enhanced baseline, the complete GCA framework generally achieves better performance across different datasets. This observation suggests that the \textcolor{black}{observed gains cannot be attributed solely to the availability of LLM-generated explanations. Instead, the combination of adaptive gating and contrastive alignment provides additional benefits beyond direct explanation fusion, although their individual effects cannot be completely separated due to the dependency between $g$ and $L_{con}$.}

\section{Discussion}
\label{sec:5}

\begin{figure*}[!t]
  \centering
  \includegraphics[width=.95\textwidth]{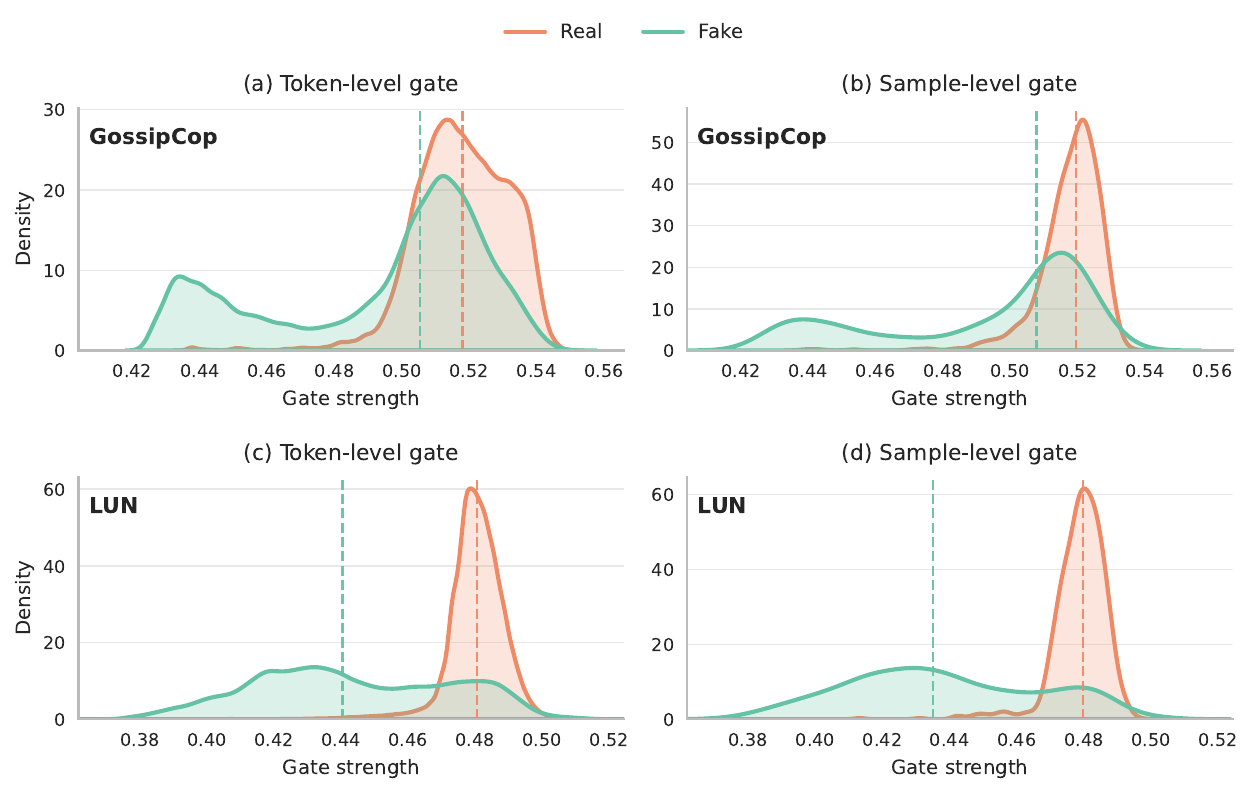}
  \caption{Gating values for real and fake news on GossipCop (a, b) and LUN (c, d) test sets. The left column shows token-level gating distributions, while the right column presents the mean gating value for each news sample. }
  \label{fig.gate}
\end{figure*}

\subsection{Gating Strength}
To further examine the gating mechanism, we visualize the distribution of gating values on the \textbf{GossipCop} and \textbf{LUN} test sets in Fig.~\ref{fig.gate}. \textcolor{black}{Both datasets contain larger test sets and provide more representative distributions, whereas the PolitiFact test set contains only 90 samples.} 

For \textbf{real news}, gating values tend to concentrate in relatively higher ranges, suggesting that the model often assigns greater importance to explanation-based contextual information. In contrast, for \textbf{fake news}, gating values are distributed over a wider range, indicating greater variation in the contribution of explanation information across samples. This behavior is consistent with our design motivation that explanation signals may be more or less informative depending on their relevance to the rewritten news.

This visualization provides qualitative evidence that the gating mechanism does not apply a uniform contribution of explanation information across samples. Its effectiveness is further supported by the ablation results in Table~\ref{tab:2}, where removing the gating mechanism leads to a noticeable performance degradation. Together, these results suggest that adaptive gating contributes to the effective integration of explanation information under emotional rewriting.

\subsection{Limitations}
This work focuses on a controlled fact-preserving emotional rewriting setting to analyze the robustness of fake news detection models under different emotional backgrounds. The main assumption in this setting is that emotionally rewritten articles inherit the authenticity labels of their original articles. The main consideration here is that manually assigning authenticity labels to a large number of rewritten samples is difficult to scale in practice. Although this assumption enables large-scale evaluation without requiring manual annotation for every generated variant, it may not fully reflect real-world scenarios, where emotional reframing can occasionally be accompanied by factual modifications.

In addition, the emotional rewritings in this study are generated by GPT-4o, which may introduce model-specific linguistic patterns. \textcolor{black}{Moreover, LLM generation introduces additional token consumption and computational overhead, which may limit the practicality of our framework in large-scale or latency-sensitive cases; human-authored alternatives, while potentially more diverse, would also require substantial effort and face scalability challenges. Future work will explore multi-LLM and human-authored rewritings, as well as settings with simultaneous changes in expressive style and factual content.} 

\subsection{Future Work}
Several directions deserve further investigation. First, constructing larger and more diverse benchmarks with human-authored or human-verified emotional rewritings would provide a more realistic robustness evaluation schema. Second, richer forms of contextual knowledge, such as retrieval-based external evidence or structured knowledge bases, may provide complementary information beyond LLM-generated explanations. Third, future studies may consider more realistic scenarios where emotional reframing is accompanied by partial factual modifications, allowing robustness under both expressive and semantic perturbations to be investigated. Finally, we plan to extend the proposed framework to multimodal fake news detection under both textual and visual perturbations.

\section{Conclusion}
\label{sec:6}
\textcolor{black}{In this work, we investigated fake news detection under fact-preserving emotional rewritings, a controlled textual setting for evaluating robustness against expressive variations. We proposed a Gated Cross Attention (GCA) framework that incorporates explanation-based contextual priors. Experiments showed improvements on PolitiFact and LUN while maintaining competitive performance on GossipCop. These results support the effectiveness of our approach in the controlled textual emotional rewriting setting. Future work will look into richer contextual knowledge and multimodal perturbations.}



\bibliographystyle{IEEEtran}
\bibliography{ref.bib}

\end{document}